\documentclass[sigconf,screen]{acmart}

\AtBeginDocument{%
  }

\usepackage{amsmath,bm}
\usepackage{algorithmic}
\usepackage[linesnumbered,ruled]{algorithm2e}
\usepackage{colortbl}
\usepackage{multirow}
\usepackage{utfsym}
\usepackage{caption}
\usepackage{subcaption}

\usepackage{array, booktabs, graphicx, tabularx, makecell}
\usepackage{setspace}
\usepackage{mathtools}
\usepackage{pifont}
\newcommand{\cmark}{\textcolor{black}{\ding{51}}}%
\newcommand{\xmark}{\textcolor{magenta}{\ding{55}}}%
\let\oldding\ding
\renewcommand{\ding}[2][1]{\scalebox{#1}{\oldding{#2}}}
\usepackage{mathrsfs}

\let\emptyset\varnothing

\iftrue

\usepackage{titlesec}
\titlespacing\section{0pt}{3pt plus 1pt minus 1pt}{0pt plus 1pt minus 1pt}
\titlespacing\subsection{0pt}{3pt plus 1pt minus 1pt}{0pt plus 1pt minus 1pt}
\titlespacing\subsubsection{0pt}{3pt plus 1pt minus 1pt}{2pt plus 1pt minus 1pt}
\fi

\setcopyright{acmcopyright}
\copyrightyear{2024}
\acmYear{2024}
\setcopyright{acmlicensed}\acmConference[DAC '24]{61st ACM/IEEE Design Automation Conference}{June 23--27, 2024}{San Francisco, CA, USA}
\acmBooktitle{61st ACM/IEEE Design Automation Conference (DAC '24), June 23--27, 2024, San Francisco, CA, USA}
\acmDOI{10.1145/3649329.3655938}
\acmISBN{979-8-4007-0601-1/24/06}
\setcopyright{none}

\begin{document}

\title{Graph Neural Networks Automated Design and Deployment \\ on Device-Edge Co-Inference Systems}

\titlenote{
This work was supported in part by National Natural Science Foundation of China (Grant No. 62072019) and National Key Laboratory of Spintronics.
Corresponding authors are \textit{Jianlei Yang} and \textit{Chunming Hu}, Email: \url{jianlei@buaa.edu.cn}, \url{hucm@buaa.edu.cn}
}

\author{Ao Zhou$^1$, \quad Jianlei Yang$^1$, \quad Tong Qiao$^1$, \quad Yingjie Qi$^1$, \quad Zhi Yang$^2$, \\ \quad Weisheng Zhao$^1$, \quad Chunming Hu$^1$} 

\affiliation{
    \institution{
        $^1$Beihang University, Beijing, China \hspace{3em}
        $^2$Peking University, Beijing, China \\
    }
    \country{}
}

\begin{abstract}

The key to device-edge co-inference paradigm is to partition models into computation-friendly and computation-intensive parts across the device and the edge, respectively.
However, for Graph Neural Networks (GNNs), we find that simply partitioning without altering their structures can hardly achieve the full potential of the co-inference paradigm due to various computational-communication overheads of GNN operations over heterogeneous devices.
We present GCoDE, the first automatic framework for \underline{G}NN that innovatively \underline{Co}-designs the architecture search and the mapping of each operation on \underline{D}evice-\underline{E}dge hierarchies.
GCoDE abstracts the device communication process into an explicit operation and fuses the search of architecture and the operations mapping in a unified space for joint-optimization. 
Also, the performance-awareness approach, utilized in the constraint-based search process of GCoDE, enables effective evaluation of architecture efficiency in diverse heterogeneous systems.
We implement the co-inference engine and runtime dispatcher in GCoDE to enhance the deployment efficiency.
Experimental results show that GCoDE can achieve up to $44.9\times$ speedup and $98.2\%$ energy reduction compared to existing approaches across various applications and system configurations.

\end{abstract}


\maketitle
\pagestyle{plain}


\section{Introduction}

Graph Neural Networks (GNNs) have emerged as the state-of-the-art (SOTA) method for graph-based learning tasks in edge scenarios such as point cloud processing \cite{li2021towards} and natural language processing \cite{wei2023neural}.
Additionally, the rising popularity of various sensors in mobile devices also encourages the deployment of GNNs to wireless network edge for tasks like sensing and interaction (e.g., collision prediction in self-driving vehicles \cite{yu2021scene}, speech analytic \cite{chen2023multivariate}).
However, GNNs often suffer from prohibitive inference costs due to their hungry demands for computational and memory resources \cite{zhang2021g}.
For example, even the optimal GNNs searched by HGNAS \cite{10247875} for edge devices only achieve about $\textbf{2}$ fps processing point cloud data on Raspberry Pi 3B+, insufficient for real-time needs.
Deploying such expensive GNNs on resource-constrained edge devices results in excessive workloads, low efficiency, and severe energy consumption, thus limiting their potential.

With the wide application of AI in real life, device-edge co-inference has emerged as a promising paradigm for the large-scale deployment of DNN models at the wireless network edge \cite{li2023roulette}.
By extracting and transmitting intermediate data from the device (e.g., smartphones) to the edge (e.g., edge nodes) for collaborative computation, this approach significantly reduces resource consumption on the device and improves efficiency.
Maximizing the benefits of co-inference depends on finding an optimal partitioning point that balances communication and computation trade-offs.
In contrast to DNNs, GNNs involve both computation-intensive matrix operations and memory-intensive graph processing \cite{zhang2021g}.
Thus, simply selecting partitioning points on an existing architecture does not fully exploit the collaborative potential.
For example, BRANCHY-GNN \cite{shao2021branchy} examines the optimal partition point for GNNs, while reducing communication overhead, does not markedly improve inference efficiency.
Therefore, an elegant approach for joint-optimization of architecture design and mapping schemes is warranted.

\begin{figure}[t]
    \centering
    \includegraphics[width = 1\linewidth]{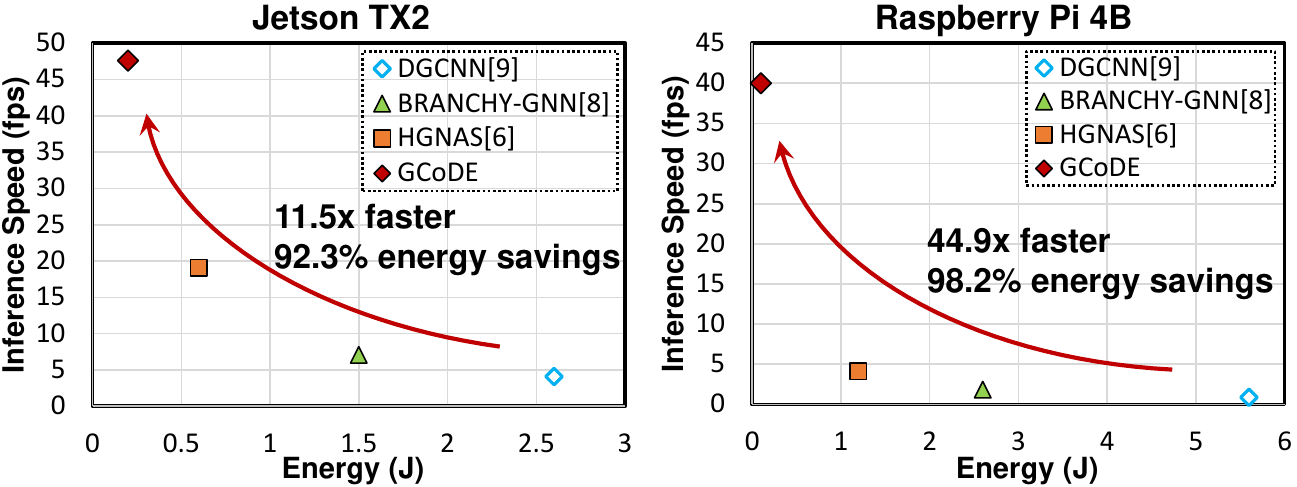}
    \caption{Inference speed vs. energy consumption comparison.}
    \label{fig:first_compare}
    \vspace{-6pt}
\end{figure}

\begin{sloppypar}
In this paper, we propose a novel NAS-based GNN architecture-mapping co-design and deployment framework for device-edge hierarchies, namely GCoDE. 
Given user requirements, GCoDE can efficiently search and deploy optimal GNN architectures with their concomitant mapping schemes for target systems, achieving both accuracy and efficiency under latency and energy constraints. 
In practice, the design and deployment of custom GNNs for device-edge hierarchies face several challenges.
First, designing GNNs for device-edge hierarchies requires balancing the trade-offs between communication and computation.
Additionally, the heterogeneity between device and edge leads to diverse GNN execution characteristics, necessitating an effective system performance awareness approach.
Moreover, separating architecture design from mapping often results in sub-optimal performance, requiring a co-optimized approach for both.
Efficient task dispatching and execution engines are also crucial for maximizing the potential of device-edge deployments.
Besides, NAS itself is known for lengthy search times.
\end{sloppypar}


To address the aforementioned challenges, our proposed GCoDE framework incorporates device-edge operation mapping into the GNN architecture design for co-optimization.
Building on a novel concept, we conceptualize the communication process between the device and edge as a specialized GNN operation. 
Such a concept is put into good use by \textbf{introducing \textit{communicate} as an explicit operation within architecture design space.}
The location of each \textit{communicate} operation in the architecture signifies a model split and the subsequent operations mapping between the device and edge.
Consequently, the mapping scheme is inherently included in the architectures sampled by GCoDE, and the exploration of the GNN architecture space will also optimize the mapping scheme.
Moreover, with the fused architecture space, system performance evaluation becomes a simpler process of architecture evaluation, thus reducing complexity.
GCoDE employs two system performance evaluation methods to guide the exploration towards more efficient design configurations.
The first approach, based on cost estimation, approximates latency relationships between architectures with low overhead.
The second, a GIN-based performance predictor powered by a feature enhancement strategy, provides accurate system latency assessments, ideal for scenarios with strict latency constraints.
By leveraging a constraint-based random search strategy, the search process requires only $3$ GPU hours.
Additionally, GCoDE integrates an efficient co-inference engine, enabling automated GNN deployment and dynamic runtime dispatch.
Fig.~\ref{fig:first_compare} shows the significant improvements in efficiency and energy savings of GCoDE over existing methods, with Intel i7 and Nvidia 1060 as edge.
The main contributions of this paper are as follows:

\begin{sloppypar}
\begin{itemize}
    \item \textbf{Framework.} To the best of our knowledge, GCoDE is the first framework for automated GNN design and deployment, targeting device-edge co-inference systems. Additionally, GCoDE achieves the co-optimization of architecture and its mapping scheme, balancing objectives like accuracy, latency, and energy.    
    \item \textbf{System performance awareness.} To our best understanding, GCoDE is also the first work to achieve system performance awareness for GNNs in heterogeneous wireless edges. The proposed system performance predictor shows over $94.7\%$ accuracy in relative latency relationship prediction among GNN architectures across diverse systems. 
    \item \textbf{Evaluation.} Extensive experiments across various applications and deployment systems highlight the superiority of GCoDE, achieving up to $44.9\times$ speedup and $98.2\%$ energy savings without sacrificing accuracy.
\end{itemize}

\end{sloppypar}


\section{Motivation and Related Works}\label{sec:motivation}
This section outlines our three key motivations based on various observations and related research.  

\begin{figure}[t]
    \centering
    \includegraphics[width = 1\linewidth]{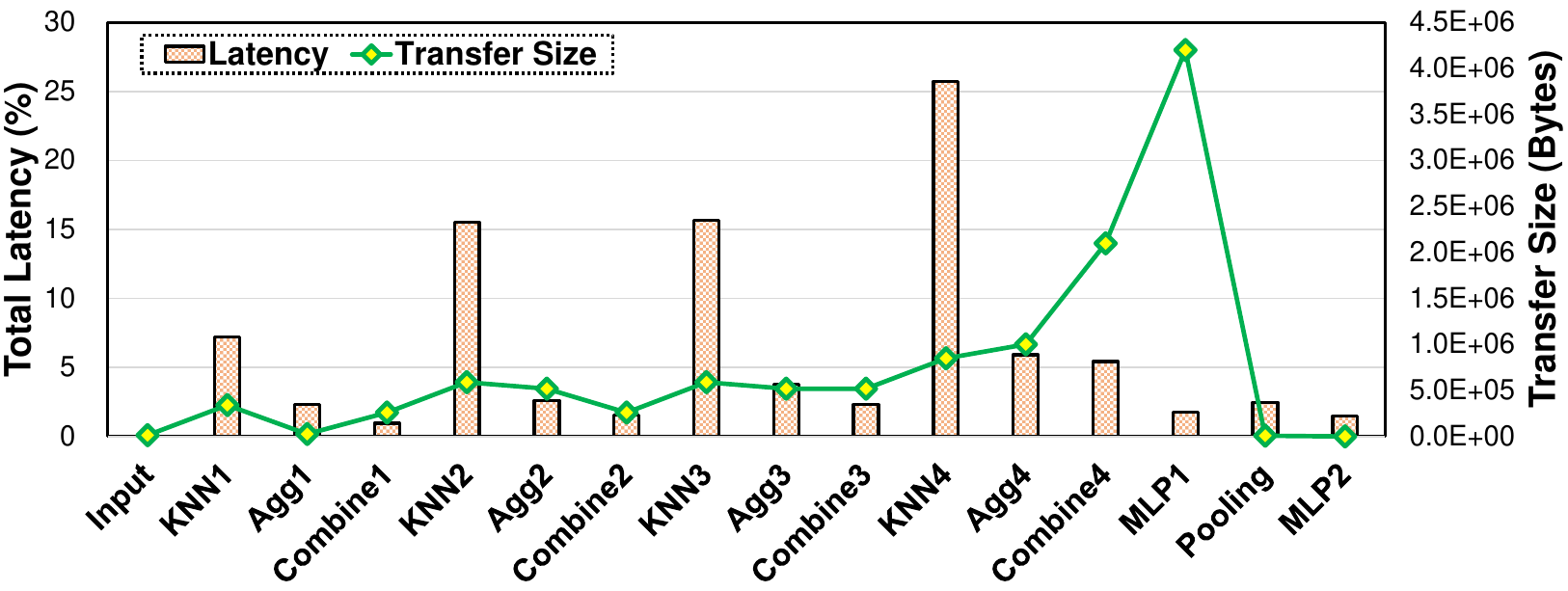}
    \caption{Changes in the required transfer data size and percentage of total latency for each operation in DGCNN.}
    \label{fig:mt1}
    \vspace{-3pt}
\end{figure}

\textbf{Motivation \raisebox{-0.1em}{\ding[1.3]{182}}: Exploring trade-offs between computation and communication on device-edge co-inference systems.}\label{sec:mt1}

Fig.~\ref{fig:mt1} illustrates the variation in computation latency for each operation in the DGCNN \cite{wang2019dynamic} architecture on ModelNet40 \cite{wu20153d} dataset and Jetson TX2 platform, alongside the data transfer sizes required for partitioning after each operation. 
As the node feature dimensions expand with each GNN layer, the \textit{KNN} operations become progressively more time-consuming, with the last one taking up over a quarter of the total execution time, aligning with findings in \cite{10247875}.
In addition, the \textit{KNN} operation produces graph data, leading to an increase in transfer size if a subsequent \textit{Aggregate} operation is required.
On the other hand, the \textit{Pooling} operation markedly reduces the intermediate data to only $68\%$ of input size, while the higher-dimensional \textit{MLP1} results in a significant increment.
As pointed out in \cite{shao2021branchy}, a potential resource-efficient deployment solution is to split the DGCNN architecture at the front layers or \textit{Pooling} operation to reduce the communication overhead. 
The effectiveness of this solution depends on the device and edge performance.
Therefore, it is necessary to carefully balance the trade-offs between communication and computation overhead during the architecture design to achieve optimal efficiency \cite{odema2021lens}.

\begin{figure}[t]
    \centering
    \includegraphics[width = 0.95\linewidth]{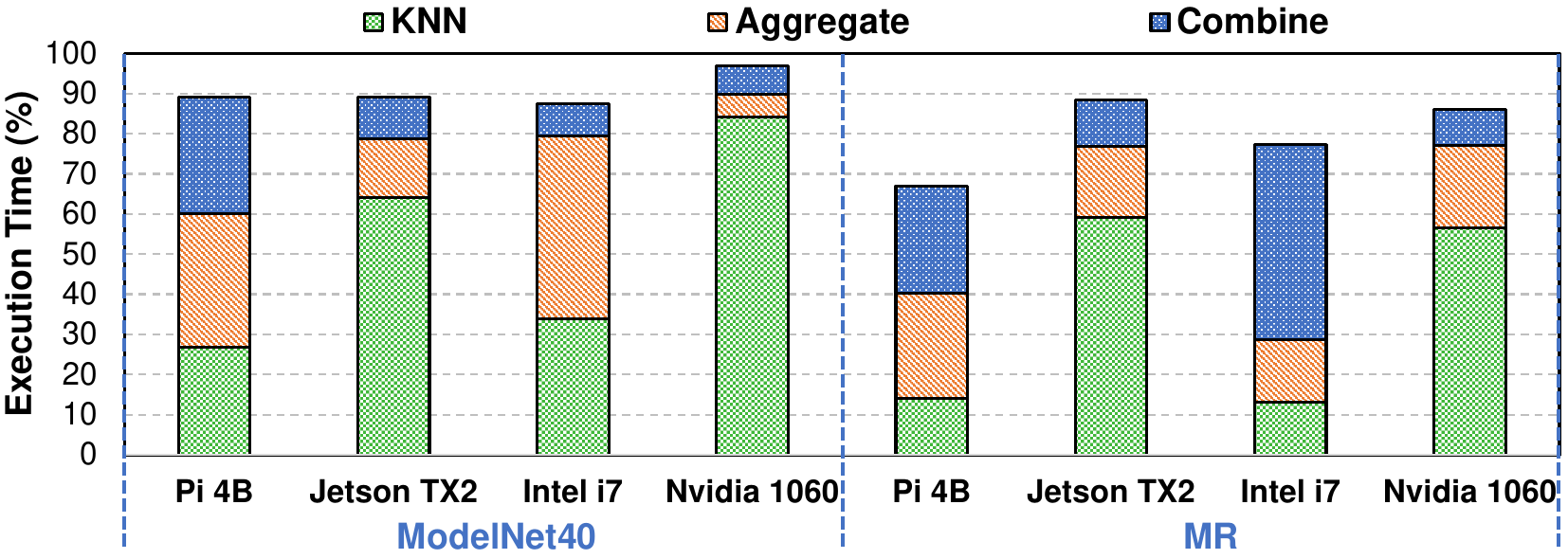}
    \caption{Execution time breakdown of DGCNN across various devices on ModelNet40 and MR datasets.}
    \label{fig:mt2}
    \vspace{-6pt}
\end{figure}

%
\textbf{Motivation \raisebox{-0.1em}{\ding[1.3]{183}}: Perceiving the heterogeneous hardware sensitivities of GNNs.}
\begin{sloppypar}
Next, we analyze the GNN hardware sensitivity by examining the execution time of DGCNN across various edge platforms.
Fig. \ref{fig:mt2} demonstrates that on the ModelNet40 point cloud dataset \cite{wu20153d}, the \textit{KNN} operation consumes most of the execution time on both Jetson TX2 and Nvidia 1060.
This is due to the intensive and irregular memory accesses, which obscure the parallel processing capabilities of GPUs.
For the Intel i7, the \textit{Aggregate} operation emerges as the main optimization challenge.
For the Raspberry Pi, constraints by the lower computing power, all operations are time-consuming.
Unlike point cloud data, the text dataset MR features fewer nodes ($1024$ vs. $17$) and larger feature dimensions ($3$ vs. $300$), resulting in distinct execution characteristics.
For instance, the \textit{Combine} operation dominates the execution time on the Intel i7.

The above observation highlights the hardware sensitivity of GNNs, emphasizing the need for performance consideration in architecture design.
HGNAS \cite{10247875} presents a GCN-based latency predictor, accurate for single-device evaluations but less effective in heterogeneous environments (see Sec.~\ref{sec:predictorResult}).
MaGNAS \cite{odema2023magnas} employs a lookup table (LUT) approach for heterogeneous MPSoCs, but this method fails to address runtime overheads \cite{benmeziane2021comprehensive}.
As a result, there remains a deficiency in elegant, scalable, and accurate approaches for assessing GNN co-inference efficiency across device-edge hierarchies.
\end{sloppypar}

\begin{figure}[t]
    \centering
    \includegraphics[width = 1\linewidth]{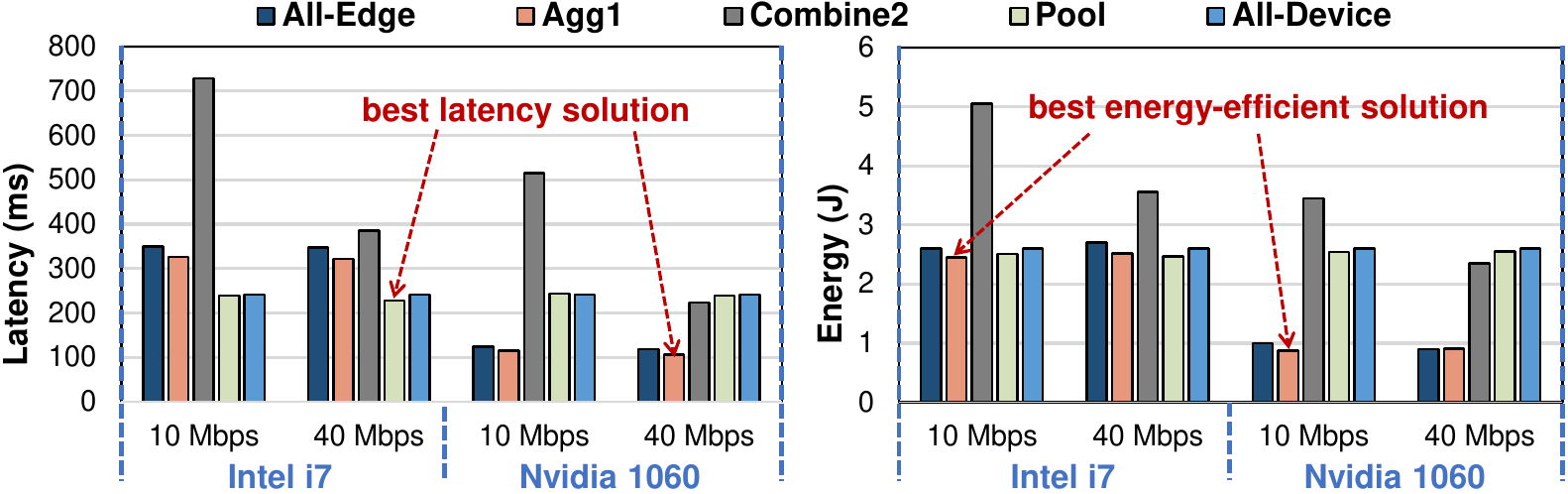}
    \caption{Performance of various partitioning schemes on DGCNN under different heterogeneities. Jetson TX2 serves as the device.}
    \label{fig:mt3}
    \vspace{-6pt}
\end{figure}


\textbf{Motivation \raisebox{-0.1em}{\ding[1.3]{184}}: Addressing the gap in automated design and deployment of GNNs on device-edge hierarchies.}

Despite significant advances in DNN co-inference research \cite{odema2021lens}, studies focusing on GNNs remain scarce.
As shown in Tab.~\ref{tab:comparison}, BRANCHY-GNN \cite{shao2021branchy} investigates the splitting and intermediate data compression techniques for GNN. 
However, the insufficient exploration of novel architectures and hardware-aware strategies leads to sub-optimal performance.
While HGANS \cite{10247875} and MaGNAS \cite{odema2023magnas} introduce hardware-aware NAS for GNNs, they lack consideration of the system heterogeneity and wireless network conditions.
A key issue is that detaching the architecture design and mapping, such as partitioning after design, the full potential of collaboration will not be realized.
Fig.~\ref{fig:mt3} shows that despite exploring potential partitioning schemes described in Motivation \raisebox{-0.1em}{\ding[1.3]{182}}, even the best partitioning points fail to significantly enhance performance under different wireless network conditions.
As such, an automated architecture-mapping co-design and deployment framework tailored for GNNs on device-edge co-inference systems is needed, which motivates our investigation in this paper.

\begin{table}[t]
\centering
\caption{Comparison of support features}
\renewcommand\arraystretch{1.1}
\resizebox{1.0\linewidth}{!}{%
\begin{tabular}{|l|c|c|c|c|}
\hline
\textbf{Supported Features} & \textbf{GCoDE} & \textbf{HGNAS \cite{10247875}} & \textbf{MaGNAS \cite{odema2023magnas}} & \textbf{BRANCHY \cite{shao2021branchy}} \\
\hline
Design Automation &   \cmark   & \cmark & \cmark & \xmark \\
\hline
Architecture Exploration & \cmark & \cmark & \cmark & \xmark \\
\hline
Performance Awareness & \cmark & \cmark & \cmark & \xmark \\
\hline
\quad $\rhd$ Single Device & \cmark & \cmark & \xmark & \xmark \\
\hline
\quad $\rhd$ Heterogeneous & \cmark & \xmark & \cmark & \xmark \\
\hline
\quad $\rhd$ Heterog. Wireless Edge & \cmark & \xmark & \xmark & \xmark \\
\hline
Multi-Objective Optimization & \cmark & \cmark & \cmark & \xmark \\
\hline
Device-Edge Deployment & \cmark & \xmark & \xmark & \cmark \\
\hline
Runtime Optimization & \cmark & \xmark & \xmark & \xmark \\
\hline
\end{tabular}
}

\label{tab:comparison}
\end{table}

\section{GCoDE Methodology}
\subsection{Framework Overview}
\begin{figure}[t]
    \centering
    \includegraphics[width = 1.0\linewidth]{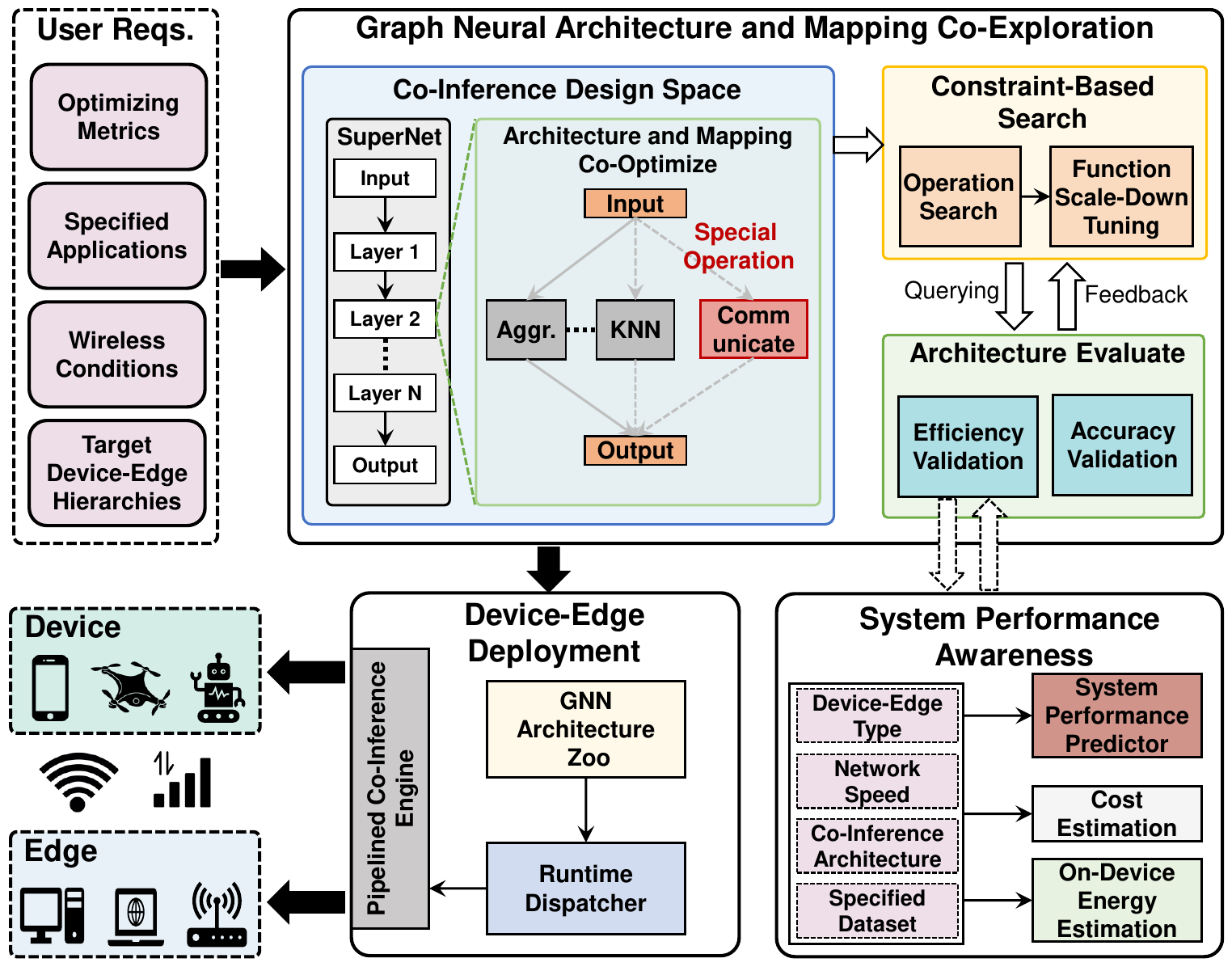}
    \caption{Overview of GCoDE framework.}
    \label{fig:framework}
    \vspace{-6pt}
\end{figure}

\begin{sloppypar}
Fig.~\ref{fig:framework} provides an overview of our GCoDE framework, comprising three key components: graph neural architecture and mapping co-exploration, system performance awareness, and device-edge deployment.
Given user requirements such as system configurations, GCoDE begins by exploring the GNN co-inference design space to locate optimal architectures with tailored mapping schemes.
GCoDE organizes the co-inference design space into a supernet, decoupling the training and searching processes via a one-shot approach.
Subsequently, GCoDE adopts a constraint-based random search strategy to efficiently explore the design space.
During the exploration, each candidate is evaluated based on its accuracy on the validation dataset and its performance (latency and energy consumption) on the target system, ensuring the user requirements for deployment.
System performance metrics are derived from our system performance predictor, cost estimation, and energy estimation method, obviating the need for laborious real-time measurements.
Following architecture exploration, a set of optimal GNN architectures is prepared for deployment. 
GCoDE assembles these architectures into an architecture zoo, enabling dynamic adjustments to the deployed architecture via our runtime dispatcher.
Deployment leverages our pipelined co-inference engine, which efficiently executes GNN inference tasks between device and edge. 
Subsequently, we will formulate the optimization process of GCoDE, and introduce these components in detail.
\end{sloppypar}

\subsection{Problem Formulation}

In this work, we aim to co-optimize the accuracy and efficiency of GNN architectures deployed on device-edge co-inference systems.
Given the user requirements: the device $\mathcal D$, the edge $\mathcal E$, anticipated network speed $\mathcal S$, latency constraint $\mathcal{C}_{lat}$ and on-device energy constraint $\mathcal{C}_{e}$, the co-optimization process can be formulated as:
\begin{equation*}
 \arg \mathop {\max }\limits_{{\alpha \in \mathbb{A}} }  \left\{ {acc_{val}}\left({{\mathcal W}^ * }, \alpha \right) - \lambda \times \left({\mathcal{P}_{sys}} \left({\alpha}, \mathcal{D}, \mathcal{E}, \mathcal{S} \right) + E_{dev}(\alpha,\mathcal{D}, \mathcal{S}) \right)\right\},
\end{equation*}
\begin{equation*}
    \begin{split}
            s.t. \quad & {{\mathcal W} ^ * } = \arg \mathop {\max }\limits_{\mathcal W}\left\{{acc_{train}} \left( {\mathcal W}, {\alpha} \right)\right\} \\
            & {{\mathcal{P}_{sys}} < \mathcal{C}_{lat}} \quad and \quad E_{dev} < \mathcal{C}_{e}
    \end{split} \quad,
\end{equation*}
where $\mathcal W$ denotes the model weights, $acc_{train}$ is the training accuracy, $acc_{val}$ is the validation accuracy, $\alpha$ is the selected architecture to be optimized form GNN co-inference design space $\mathbb{A}$, $\mathcal{P}_{sys}$ indicates the inference latency on targeted system,
$E_{dev}$ denotes the energy consumption of $\alpha$ on the device $\mathcal D$, and $\lambda$ is a scaling factor used to adjust the optimize propensity between accuracy and efficiency.
Note that system performance is jointly determined by the architecture, device-edge configurations, and network conditions.

\subsection{Co-Inference Design Space}

\begin{figure}[t]
    \centering
    \includegraphics[width = 1.0\linewidth]{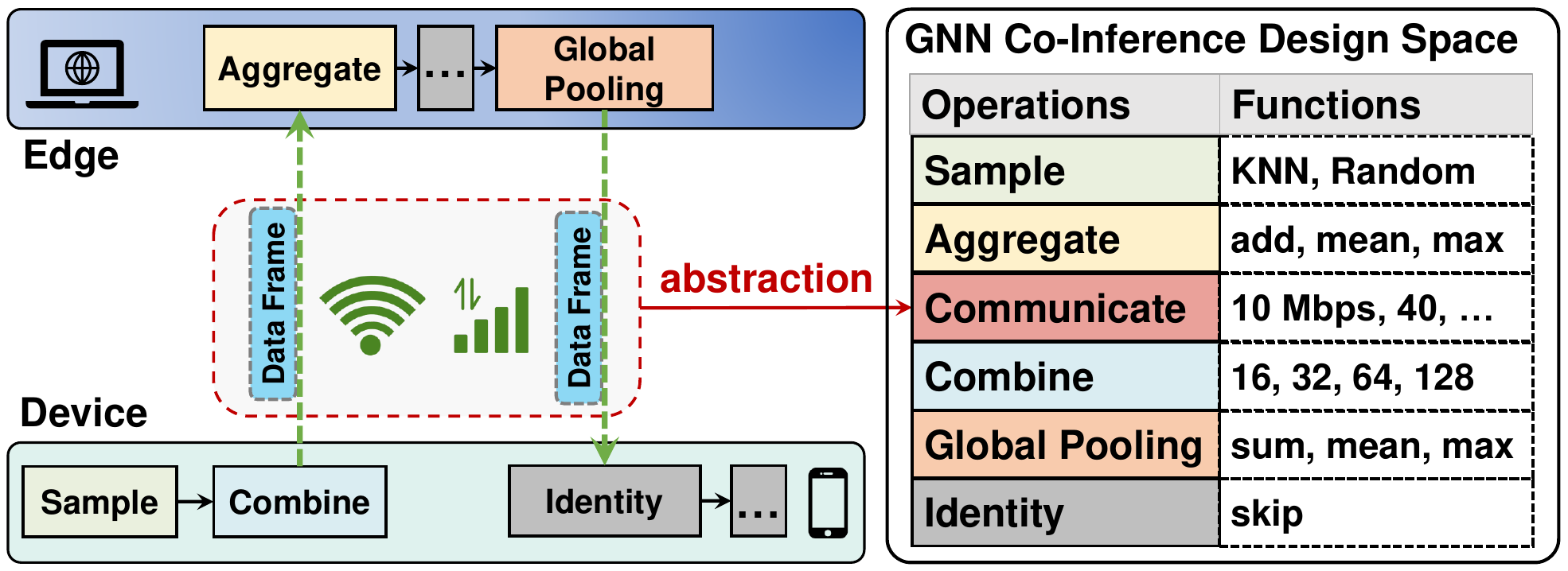}
    \caption{Unified design space of architecture design and mapping.}
    \label{fig:mapping}
    \vspace{-6pt}
\end{figure}

Fig.~\ref{fig:mapping} illustrates our approach to avoid the detachment of architecture design and mapping by establishing a unified co-inference design space for GNN.
Specifically, the primary difference between co-inference and single-device inference lies in the data communication between device and edge, making the selection of an effective communication point vital.
To achieve an optimal balance between communication and computation, we introduce a novel concept: communication between device and edge can be treated as a specialized operation within the GNN architecture.
Thus, we abstract device communication as a distinct GNN operation and integrate it into the architecture design space.
This fusion of architecture and mapping enables GCoDE to explore various flexible collaborative patterns, rather than simply searching for a partition point.
Furthermore, this flexible mapping aids in adjusting to system heterogeneity and discovering optimal collaboration patterns during exploration.

Specifically, the GNN co-inference design space $\mathbb{A}$ is organized as a supernet, comprising six operations in each layer: \textit{Sample}, \textit{Aggregate}, \textit{Communicate}, \textit{Combine}, \textit{Global Pooling}, and \textit{Identity}, each having distinct functional properties as shown in Fig.~\ref{fig:mapping}.
During supernet training, linear layers are used to align the dimensions of all operations within the same layer, which will be removed before search to ensure efficiency. 

\begin{algorithm}[t]
\small 
\setstretch{0.9} 
\KwIn{Target system ${Sys}$, constraints: $\{\mathcal{C}_{lat}, \mathcal{C}_{e}\}$, function setting $f$, max search and tuning iteration:{ $T$, $T_{f}$}.}
\KwOut{The best found GNN architectures $\alpha^*$.}
Initialize $\alpha^* \gets \emptyset$, $Ops \gets \emptyset$, $func \gets \emptyset$.\\
Pre-train GNN supernet $\mathcal{N}_{super}$ with $f$.\\   
{\textcolor{blue}{/* Stage 1: operation search */}} \\
\For{$1 \leq t \leq T$}{
    \While{\textnormal{Check($Ops$)}}{
        $Ops \gets \textnormal{Random}\left(\mathbb{A}\right)$ \hfill{\textcolor{brown}{// Sample valid operations}}\\
    }
    $\left\{\mathcal{P}_{sys}, E_{dev}\right\} \gets \textnormal{Evaluate}\left(Sys, Ops, f\right)$ \hfill{\textcolor{brown}{// Evaluate performance}}\\
    \eIf{$\mathcal{P}_{sys} < \mathcal{C}_{lat}$ \textnormal{and} $E_{dev} < \mathcal{C}_{e}$}{
        $score \gets \left( acc_{val} - \lambda (\mathcal{P}_{sys} + E_{dev})\right)$ \hfill{\textcolor{brown}{// Full evaluation}}
    }{$score \gets (-1)$ \hfill{\textcolor{brown}{// Discard failed architectures}}}
    $\alpha^*$ $\gets$ \textnormal{Update($Ops$, $f$, $score$)} \hfill{\textcolor{brown}{// Update operations}}\\
}
{\textcolor{blue}{/* Stage 2: function scale-down tuning */}} \\
\For{$1 \leq t \leq T_{f}$}{
$func \gets \textnormal{Random}(\mathbb{A}, f)$ \hfill{\textcolor{brown}{// scale-down functions from $f$}}\\
$\alpha^*$ $\gets$ \textnormal{Update($Ops$, $func$, $acc_{val}$)} \hfill{\textcolor{brown}{// Update functions}}\\
}
\textbf{return} $\alpha^*$ \hfill{\textcolor{brown}{// Top-performing designs}}
\caption{\small Constraint-based search strategy.}
\label{gnnalgo}
\vspace{-3pt}
\end{algorithm}

\subsection{Efficient Constraint-Based Search Strategy}


Incorporating mapping schemes into the GNN design space leads to numerous invalid candidates during exploration.
Examples of this are consecutive \textit{Communicate} operations or a \textit{Global Pooling} operation followed by \textit{Aggregate}.
In such cases, intelligent algorithms, such as evolutionary algorithms (EA), face challenges in identifying valid architectures (see Sec.~\ref{sec:ablation}).
Conversely, the simpler random search can yield surprising benefits in such complex design space \cite{yu2019evaluating}.
Furthermore, random search is more customizable and supports maintaining multiple optimal solutions for various objectives within a single search, catering to the varied application requirements of co-inference (e.g., low energy and low latency).
As a result, GCoDE adopts a constraint-based random search strategy to improve exploration efficiency, as shown in Alg.~\ref{gnnalgo}.
Given user requirements, GCoDE sets an appropriate function configuration for the supernet, guided by the target architecture like DGCNN.
Subsequently, GCoDE pre-trains the supernet with a focus on accuracy to establish shared weights for further architecture exploration.
Details of the search process are outlined below.

\textbf{Stage 1: Operation search.} This stage focuses on identifying optimal operation sets that adhere to the system performance constraints and accuracy requirements.
Specifically, GCoDE checks the validity of each randomly sampled operation set to prevent interference from invalid architectures during the search.
Only architectures that pass this validity check are subsequently evaluated for inference latency $\mathcal{P}_{sys}$ and energy consumption $E_{dev}$.
Additionally, accuracy assessments are performed only on architectures that meet the specified constraints.
To mitigate the influence of varying metrics magnitudes on optimization efficacy, $\mathcal{P}_{sys}$ and $E_{dev}$ are normalized during architecture scoring.
This combined approach of validity and performance constraint checks effectively minimizes unnecessary evaluation overheads.

\textbf{Stage 2: Function scale-down tuning.}
This stage aims to find more efficient function settings while maintaining accuracy.
Practically, architectures that pass the first stage already satisfy the performance requirements.
Therefore, GCoDE concentrates on evaluating the impact of scaled-down function settings on accuracy in this stage.
Specifically, GCoDE performs scale-down tuning of function setting $f$, such as reducing the dimensions of $Combine$, to explore more efficient architectures.
After the search, GCoDE generates a set of optimal architectures, which are maintained by the GNN architecture zoo to wait for deployment.

\subsection{System Performance Awareness}
\begin{figure}[t]
    \centering
    \includegraphics[width = 1.0\linewidth]{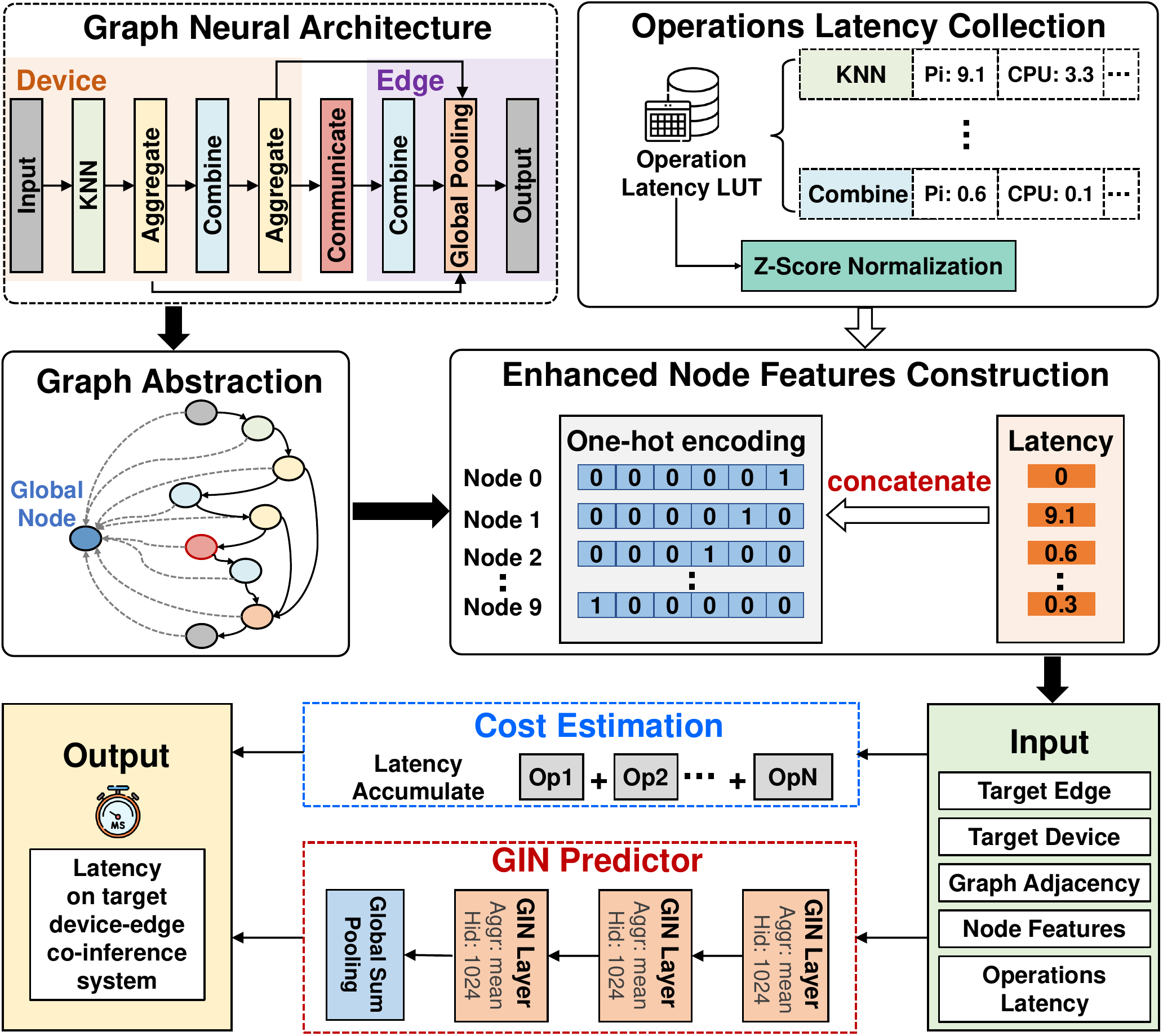}
    \caption{Latency perception for GNNs on device-edge hierarchies.}
    \label{fig:predictor}
    \vspace{-6pt}
\end{figure}
GCoDE integrates system performance awareness methods to guarantee efficiency post-deployment.
Specifically, the latency perceive process includes graph abstraction, enhanced node features construction, and inference cost estimation or latency prediction, as depicted in Fig.~\ref{fig:predictor}.

\textbf{Graph abstraction.}
Drawing inspiration from \cite{10247875}, GCoDE abstracts GNN architecture into a directed graph to facilitate architecture graph learning.
In this graph, nodes represent various operations, while edges indicate data flow between operations.
Additionally, GCoDE introduces self-connections and a global node to enhance graph connectivity.

\textbf{Enhanced node features construction.}
To better capture system heterogeneity and network conditions, GCoDE develops enhanced node features for the architecture graph.
Specifically, GCoDE maintains an operation latency LUT across various devices, with negligible construction overhead due to the limited number of valid operations.
The \textit{communicate} operation latency is calculable based on the transfer data size and the available network bandwidth.
GCoDE then concatenates the one-hot encoding of each node with its corresponding latency from the LUT, enriching the initial node features.
To mitigate the effect of varying operation magnitudes, latency values are normalized using \textit{z-score normalization} before concatenation.

\textbf{Cost estimation.}
With the mapping scheme integrated into the architecture, system performance evaluation becomes efficient and straightforward.
Specifically, based on the maintained latency LUT, we can easily accumulate all operation latency in the architecture graph.
While this estimation may not include potential runtime overheads compared to measured latency, it effectively captures the relative latency relationship between architectures, crucial for steering the exploration process towards more efficient designs.
In application scenarios without strict latency constraints, this scheme is highly cost-effective.

\textbf{Latency prediction.}
By abstracting the architecture into a graph, system performance evaluation is transformed into a graph learning problem, an area where GNNs excel.
Moreover, GIN \cite{xu2019powerful} outperforms other GNN layers in extracting comprehensive graph information.
Consequently, we built a latency predictor using three GIN layers, each with a \textit{mean} aggregation operator.
Besides, \textit{Global Sum Pooling} is used as the extractor for all node features.
The combination of \textit{mean} aggregation operator with \textit{Global Sum Pooling} enables the predictor to efficiently extract latency information from the total graph.
Due to the small number of nodes in the architecture graph, the runtime overhead of the predictor is minimal, measured in milliseconds on GPUs.
The highly accurate system latency predictor ensures that the explored architecture meets the strict latency requirements of specific application scenarios (e.g., autonomous driving).

\textbf{On-device energy estimation.}
To meet the energy constraints, we employ the energy estimation method described in \cite{odema2021lens}, focusing on the energy consumption of the device.
Specifically, the total energy consumption of the device for a single inference is estimated as: 
\begin{equation*}
    E_{total} = E_{idle} + E_{run} + E_{comm}\ ,
\end{equation*}
where $E_{comm}$ represents the communication energy consumption, computed using power models proposed in \cite{huang2012close}. 
$E_{idle}$ and $E_{run}$ are computational energy consumption in idle and operation executed states, respectively, calculated by multiplying associated power consumption with idle and execution time.

\subsection{Device-Edge Deployment}
The deployment of GCoDE is based on the integrated architecture zoo, runtime dispatcher, and co-inference engine, as listed in Fig.~\ref{fig:framework}.

\textbf{GNN architecture zoo.} To accommodate diverse runtime requirements, GCoDE maintains a set of optimal GNN co-inference architectures (low energy consumption, low latency, high accuracy, etc.) in an architecture zoo. 
With the proposed constraint-based search strategy, GCoDE generates all these GNNs in a single search without additional overheads.

\textbf{Runtime dispatcher.}
Furthermore, GCoDE dynamically adapts execution architectures via its runtime dispatcher to meet the fluctuating latency and power consumption constraints of the device.
This flexibility stems from the GNN architecture zoo, ensuring optimal architecture selection for varied environments.

\textbf{Co-inference engine.} We develop an efficient co-inference engine using Python socket \cite{socket}, enabling pipelined task execution.
Specifically, instead of waiting for the edge to return execution results after processing a data frame, the device immediately begins processing the next frame.
This approach effectively mitigates communication latency, significantly boosting the overall processing efficiency of the co-inference application.
Additionally, sending and receiving processes are implemented on separate threads. Each thread maintains its own message queue, allowing for better handling of diverse network conditions.
To reduce communication overhead, GCoDE compresses all transmitted data based on zlib tool \cite{zlib}.


\begin{table*}[t]
\centering
\renewcommand\arraystretch{1}
\caption{Performance comparison of GCoDE and existing approaches in different modes: Device-Only (D), Edge-Only (E), and Device-Edge Co-Inference (Co). OA and mAcc denote overall and balanced accuracy, respectively.}
\resizebox{\linewidth}{!}{%
\begin{tabular}{|c|c|c|c|c|cc|cc|cc|cc|}
\hline
\multirow{2}{*}{\textbf{$\mathcal{S}_{L}$}} &
  \multirow{2}{*}{\textbf{Method}} &
  \multirow{2}{*}{\textbf{OA (\%)}} &
  \multirow{2}{*}{\textbf{mAcc (\%)}} &
  \multirow{2}{*}{\textbf{Mode}} &
  \multicolumn{2}{c|}{\textbf{Jetson TX2 $\rightleftharpoons$ Nvidia 1060}} &
  \multicolumn{2}{c|}{\textbf{Jetson TX2 $\rightleftharpoons$ Intel i7}} &
  \multicolumn{2}{c|}{\textbf{Pi 4B $\rightleftharpoons$ Nvidia 1060}} &
  \multicolumn{2}{c|}{\textbf{Pi 4B $\rightleftharpoons$ Intel i7}} \\ \cline{6-13} 
   &
   &
   &
   &
   &
  \multicolumn{1}{c|}{\textbf{Latency (ms)}} &
  \textbf{Energy (J)} &
  \multicolumn{1}{c|}{\textbf{Latency (ms)}} &
  \textbf{Energy (J)} &
  \multicolumn{1}{c|}{\textbf{Latency (ms)}} &
  \textbf{Energy (J)} &
  \multicolumn{1}{c|}{\textbf{Latency (ms)}} &
  \textbf{Energy (J)} \\ \hline
\multirow{9}{*}{\rotatebox{90}{\textbf{$\le$ 40 Mbps}}} &
  \multirow{2}{*}{DGCNN \cite{wang2019dynamic}} &
  \multirow{2}{*}{92.9} &
  \multirow{2}{*}{88.9} &
  D &
  \multicolumn{1}{c|}{241.9} &
  2.6 &
  \multicolumn{1}{c|}{241.9} &
  2.6 &
  \multicolumn{1}{c|}{1121.8} &
  5.6 &
  \multicolumn{1}{c|}{1121.8} &
  5.6 \\[-2pt]
 &
   &
   &
   &
  E &
  \multicolumn{1}{c|}{118.8 (2.0$\times$$\uparrow$)} &
  0.9 (65.4\%$\downarrow$) &
  \multicolumn{1}{c|}{347.6 (1.4$\times$$\downarrow$)} &
  2.6 (0.0\%$\downarrow$) &
  \multicolumn{1}{c|}{103.5 (10.8$\times$$\uparrow$)} &
  0.4 (92.9\%$\downarrow$) &
  \multicolumn{1}{c|}{333.7 (3.4$\times$$\uparrow$)} &
  1.0 (82.1\%$\downarrow$) \\[2pt]
 
 &
  \multirow{2}{*}{\cite{li2021towards}} &
  \multirow{2}{*}{92.6} &
  \multirow{2}{*}{90.6} &
  D &
  \multicolumn{1}{c|}{107.6 (2.3$\times$$\uparrow$)} &
  1.2 (53.8\%$\downarrow$) &
  \multicolumn{1}{c|}{107.6 (2.2$\times$$\uparrow$)} &
  1.2 (53.8\%$\downarrow$) &
  \multicolumn{1}{c|}{851.1 (1.3$\times$$\uparrow$)} &
  4.3 (23.2\%$\downarrow$) &
  \multicolumn{1}{c|}{851.1 (1.3$\times$$\uparrow$)} &
  4.3 (23.2\%$\downarrow$) \\[-2pt]
 &
   &
   &
   &
  E &
  \multicolumn{1}{c|}{86.6 (2.8$\times$$\uparrow$)} &
  0.7 (73.1\%$\downarrow$) &
  \multicolumn{1}{c|}{321.6 (1.3$\times$$\downarrow$)} &
  2.5 (3.8\%$\downarrow$) &
  \multicolumn{1}{c|}{68.7 (16.3$\times$$\uparrow$)} &
  0.2 (96.4\%$\downarrow$) &
  \multicolumn{1}{c|}{303.4 (3.7$\times$$\uparrow$)} &
  1.0 (82.1\%$\downarrow$) \\[2pt]
 
 &
  \multirow{2}{*}{HGNAS \cite{10247875}} &
  \multirow{2}{*}{92.1$\sim$92.5} &
  \multirow{2}{*}{88.3$\sim$88.8} &
  D &
  \multicolumn{1}{c|}{52.1 (4.6$\times$$\uparrow$)} &
  0.6 (76.9\%$\downarrow$) &
  \multicolumn{1}{c|}{52.1 (4.6$\times$$\uparrow$)} &
  0.6 (76.9\%$\downarrow$) &
  \multicolumn{1}{c|}{241.5 (4.6$\times$$\uparrow$)} &
  1.2 (78.6\%$\downarrow$) &
  \multicolumn{1}{c|}{241.5 (4.6$\times$$\uparrow$)} &
  1.2 (78.6\%$\downarrow$) \\[-2pt]
   &
   &
   &
   &
  E &
  \multicolumn{1}{c|}{79.2 (3.1$\times$$\uparrow$)} &
  0.6 (76.9\%$\downarrow$) &
  \multicolumn{1}{c|}{83.7 (2.9$\times$$\uparrow$)} &
  0.7 (73.1\%$\downarrow$) &
  \multicolumn{1}{c|}{69.0 (16.3$\times$$\uparrow$)} &
  0.2 (96.4\%$\downarrow$) &
  \multicolumn{1}{c|}{71.0 (15.8$\times$$\uparrow$)} &
  0.3 (94.6\%$\downarrow$) \\[2pt]
 
 &
  BRANCHY \cite{shao2021branchy} &
  92.0 &
  - &
  Co &
  \multicolumn{1}{c|}{141.2 (1.7$\times$$\uparrow$)} &
  1.5 (42.3\%$\downarrow$) &
  \multicolumn{1}{c|}{140.2 (1.7$\times$$\uparrow$)} &
  1.5 (42.3\%$\downarrow$) &
  \multicolumn{1}{c|}{541.8 (2.1$\times$$\uparrow$)} &
  2.6 (53.6\%$\downarrow$) &
  \multicolumn{1}{c|}{528.1 (2.1$\times$$\uparrow$)} &
  2.5 (55.4\%$\downarrow$) \\[2pt]
 
 &
  \makecell[c]{HGNAS \cite{10247875}+Partition} &
  92.1$\sim$92.2 &
  88.3$\sim$88.7 &
  Co &
  \multicolumn{1}{c|}{52.6 (4.6$\times$$\uparrow$)} &
  0.5 (80.8\%$\downarrow$) &
  \multicolumn{1}{c|}{51.4 (4.7$\times$$\uparrow$)} &
  0.5 (80.8\%$\downarrow$) &
  \multicolumn{1}{c|}{53.7 (20.9$\times$$\uparrow$)} &
  1.9 (66.1\%$\downarrow$) &
  \multicolumn{1}{c|}{106.0 (10.6$\times$$\uparrow$)} &
  2.1 (62.5\%$\downarrow$) \\[2pt]

 &
  \textbf{GCoDE} &
  \textbf{92.1$\sim$92.6} &
  \textbf{88.1$\sim$89.7} &
  \textbf{Co} &
  \multicolumn{1}{c|}{\textbf{31.9 (7.6$\times$$\uparrow$)}} &
  \textbf{0.3 (88.5\%$\downarrow$)} &
  \multicolumn{1}{c|}{\textbf{21.0 (11.5$\times$$\uparrow$)}} &
  \textbf{0.2 (92.3\%$\downarrow$)} &
  \multicolumn{1}{c|}{\textbf{25.0 (44.9$\times$$\uparrow$)}} &
  \textbf{0.1 (98.2\%$\downarrow$)} &
  \multicolumn{1}{c|}{\textbf{64.4 (17.4$\times$$\uparrow$)}} &
  \textbf{0.2 (96.4\%$\downarrow$)} \\ \hline 

\multirow{9}{*}{\rotatebox{90}{\textbf{$\le$ 10 Mbps}}} &
  \multirow{2}{*}{DGCNN \cite{wang2019dynamic}} &
  \multirow{2}{*}{92.9} &
  \multirow{2}{*}{88.9} &
  D &
  \multicolumn{1}{c|}{241.9} &
  2.6 &
  \multicolumn{1}{c|}{241.9} &
  2.6 &
  \multicolumn{1}{c|}{1121.8} &
  5.6 &
  \multicolumn{1}{c|}{1121.8} &
  5.6 \\[-2pt]
 &
   &
   &
   &
  E &
  \multicolumn{1}{c|}{123.9 (2.0$\times$$\uparrow$)} &
  1.0 (61.5\%$\downarrow$) &
  \multicolumn{1}{c|}{350.1 (1.4$\times$$\downarrow$)} &
  2.7 (3.8\%$\uparrow$) &
  \multicolumn{1}{c|}{107.8 (10.4$\times$$\uparrow$)} &
  0.4 (92.9\%$\downarrow$) &
  \multicolumn{1}{c|}{339.5 (3.3$\times$$\uparrow$)} &
  1.1 (80.4\%$\downarrow$) \\[2pt]
 
 &
  \multirow{2}{*}{\cite{li2021towards}} &
  \multirow{2}{*}{92.6} &
  \multirow{2}{*}{90.6} &
  D &
  \multicolumn{1}{c|}{107.6 (2.2$\times$$\uparrow$)} &
  1.2 (53.8\%$\downarrow$) &
  \multicolumn{1}{c|}{107.6 (2.2$\times$$\uparrow$)} &
  1.2 (53.8\%$\downarrow$) &
  \multicolumn{1}{c|}{851.1 (1.3$\times$$\uparrow$)} &
  4.3 (23.2\%$\downarrow$) &
  \multicolumn{1}{c|}{851.1 (1.3$\times$$\uparrow$)} &
  4.3 (23.2\%$\downarrow$) \\[-2pt]
 &
   &
   &
   &
  E &
  \multicolumn{1}{c|}{93.4 (2.6$\times$$\uparrow$)} &
  0.7 (73.1\%$\downarrow$) &
  \multicolumn{1}{c|}{325.5 (1.3$\times$$\downarrow$)} &
  2.5 (3.8\%$\downarrow$) &
  \multicolumn{1}{c|}{75.8 (14.8$\times$$\uparrow$)} &
  0.3 (94.6\%$\downarrow$) &
  \multicolumn{1}{c|}{307.7 (3.6$\times$$\uparrow$)} &
  1.0 (82.1\%$\downarrow$) \\[2pt]

 &
  \multirow{2}{*}{HGNAS \cite{10247875}} &
  \multirow{2}{*}{92.1$\sim$92.5} &
  \multirow{2}{*}{88.3$\sim$88.8} &
  D &
  \multicolumn{1}{c|}{52.1 (4.6$\times$$\uparrow$)} &
  0.6 (76.9\%$\downarrow$) &
  \multicolumn{1}{c|}{52.1 (4.6$\times$$\uparrow$)} &
  0.6 (76.9\%$\downarrow$) &
  \multicolumn{1}{c|}{241.5 (4.6$\times$$\uparrow$)} &
  1.2 (78.6\%$\downarrow$) &
  \multicolumn{1}{c|}{241.5 (4.6$\times$$\uparrow$)} &
  1.2 (78.6\%$\downarrow$) \\[-2pt]
 &
   &
   &
   &
  E &
  \multicolumn{1}{c|}{87.8 (2.8$\times$$\uparrow$)} &
  0.7 (73.1\%$\downarrow$) &
  \multicolumn{1}{c|}{88.3 (2.7$\times$$\uparrow$)} &
  0.7 (73.1\%$\downarrow$) &
  \multicolumn{1}{c|}{70.3 (16.0$\times$$\uparrow$)} &
  0.3 (94.6\%$\downarrow$) &
  \multicolumn{1}{c|}{74.0 (15.2$\times$$\uparrow$)} &
  0.3 (94.6\%$\downarrow$) \\[2pt]
 
 &
  BRANCHY \cite{shao2021branchy} &
  92.0 &
  - &
  Co &
  \multicolumn{1}{c|}{141.0 (1.7$\times$$\uparrow$)} &
  1.5 (42.3\%$\downarrow$) &
  \multicolumn{1}{c|}{140.8 (1.7$\times$$\uparrow$)} &
  1.5 (42.3\%$\downarrow$) &
  \multicolumn{1}{c|}{531.8 (2.1$\times$$\uparrow$)} &
  2.6 (53.6\%$\downarrow$) &
  \multicolumn{1}{c|}{544.0 (2.1$\times$$\uparrow$)} &
  2.6 (53.6\%$\downarrow$) \\[2pt]
 
 &
  HGNAS \cite{10247875}+Partition &
  92.1$\sim$92.2 &
  88.3$\sim$88.7 &
  Co &
  \multicolumn{1}{c|}{57.1 (4.2$\times$$\uparrow$)} &
  0.5 (80.8\%$\downarrow$) &
  \multicolumn{1}{c|}{53.8 (4.5$\times$$\uparrow$)} &
  0.5 (80.8\%$\downarrow$) &
  \multicolumn{1}{c|}{72.9 (15.4$\times$$\uparrow$)} &
  0.9 (83.9\%$\downarrow$) &
  \multicolumn{1}{c|}{122.8 (9.1$\times$$\uparrow$)} &
  1.0 (82.1\%$\downarrow$) \\[2pt]
 
 &
  \textbf{GCoDE} &
  \textbf{92.2$\sim$92.8} &
  \textbf{88.7$\sim$89.7} &
  \textbf{Co} &
  \multicolumn{1}{c|}{\textbf{39.0 (6.2$\times$$\uparrow$)}} &
  \textbf{0.3 (88.5\%$\downarrow$)} &
  \multicolumn{1}{c|}{\textbf{50.2 (4.8$\times$$\uparrow$)}} &
  \textbf{0.5 (80.8\%$\downarrow$)} &
  \multicolumn{1}{c|}{\textbf{35.6 (31.5$\times$$\uparrow$)}} &
  \textbf{0.1 (98.2\%$\downarrow$)} &
  \multicolumn{1}{c|}{\textbf{49.3 (22.8$\times$$\uparrow$)}} &
  \textbf{0.2 (96.4\%$\downarrow$)} \\ \hline
\end{tabular}%
}
\label{tab:modelnet40}
\vspace{-6pt}
\end{table*}

\section{Experiments}

\subsection{Experimental Setup}

\begin{sloppypar}
\textbf{Datasets and competitors settings.}
To evaluate GCoDE, we consider two different application datasets: the point cloud processing benchmark ModelNet40 \cite{wu20153d} and the text analysis dataset MR \cite{ZhangYCWWW20}, following the evaluation settings in \cite{10247875, wei2023neural}.
Our comparison included several baselines: (1) the manually designed DGCNN \cite{wang2019dynamic}, (2) the manually optimized architecture \cite{li2021towards}, (3) the GNN device-edge co-inference method BRANCHY-GNN (denoted as BRANCHY) \cite{shao2021branchy}, and (4) two GNN NAS framework HGNAS \cite{10247875} and PNAS \cite{wei2023neural}.
Additionally, to compare architecture-mapping separation designs and joint optimization, we partition hardware-efficient GNNs from existing NAS frameworks and evaluated the efficiency at optimal partition points.
For a fair comparison, we used the reported task accuracy in these papers and tested efficiency based on the PyTorch Geometric (PyG) framework \cite{Fey/Lenssen/2019} under the same experimental conditions, averaging results from 10 runs.
\end{sloppypar}

\textbf{Devices and implementation settings.}
To compare the efficiency of GCoDE and competitors, we employ four device-edge configurations: Jetson TX2~\cite{tx2} and Raspberry Pi 4B~\cite{raspberry} as device, and Nvidia 1060 GPU~\cite{1060} and Intel i7-7700 CPU~\cite{cpu} as edge.
All devices are connected to a wireless router, with varying network conditions simulated by setting upload bandwidth limits ($\mathcal{S}_{L}$) at $10$ Mbps and $40$ Mbps.
Furthermore, all implementations and tests are conducted on the PyG framework, ensuring test reliability.
The architecture search is conducted over a maximum of $2000$ iterations and $10$ tuning iterations.
For predictor training, we randomly sampled $9$K co-inference architectures ($70\%/30\%$ for training/validation) and trained them for $200$ epochs using MAPE as the loss function.

\subsection{Evaluation on ModelNet40}

\begin{sloppypar}
\textbf{GCoDE vs. Existing approaches.}
Tab.~\ref{tab:modelnet40} compares the task accuracy, latency, and on-device energy consumption of GCoDE against all baselines.
To demonstrate the overall enhancement from architecture-mapping co-design and performance awareness on co-inference efficiency, we also compare the performance of all baselines under various collaboration modes.
It is clear to see that the performance improvement of GCoDE in device-edge co-inference significantly surpasses existing manual designs and NAS methods without sacrificing accuracy.
In case of better network conditions of $40$ Mbps, GCoDE can achieve up to $44.9\times$ speedup compared to DGCNN.
Against hardware-efficient GNNs designed by HGNAS for edge devices, GCoDE consistently shows optimal performance, achieving up to $9.7\times$ speedup on the lower-powered Raspberry Pi.
Compared with the GNN co-inference method BRANCHY, GCoDE can also achieve up to $21.7\times$ speedup, highlighting the importance of performance-awareness in architecture design.
Compared to HGNAS with its best partitioning point selection, GCoDE still achieves up to $2.5\times$ inference speedup.
Notably, processing all data at the Edge (Edge-Only mode) often fails to achieve optimal performance, primarily because of high communication overhead and wasted computing power of the device.
In case of worse network conditions of $10$ Mbps, GCoDE maintains its superiority over all baselines by leveraging its environmental awareness, achieving up to $14.9\times$ speedup compared to BRANCHY.
Additionally, GCoDE offers substantial on-device energy savings, outperforming all baselines with up to $98.2\%$ reduction in energy consumption, enhancing the potential of GNNs on resource-limited edge devices.
The results show that the architecture and deployment mapping co-design of GCoDE allows both device and edge to achieve their full potential and optimal system performance.
Moreover, the advanced system performance awareness method of GCoDE effectively identifies optimal solutions in the device-edge co-inference paradigm, ensuring scalability across various system configurations.
\end{sloppypar}

\begin{figure}[t]
    \centering
    \includegraphics[width = 0.9\linewidth]{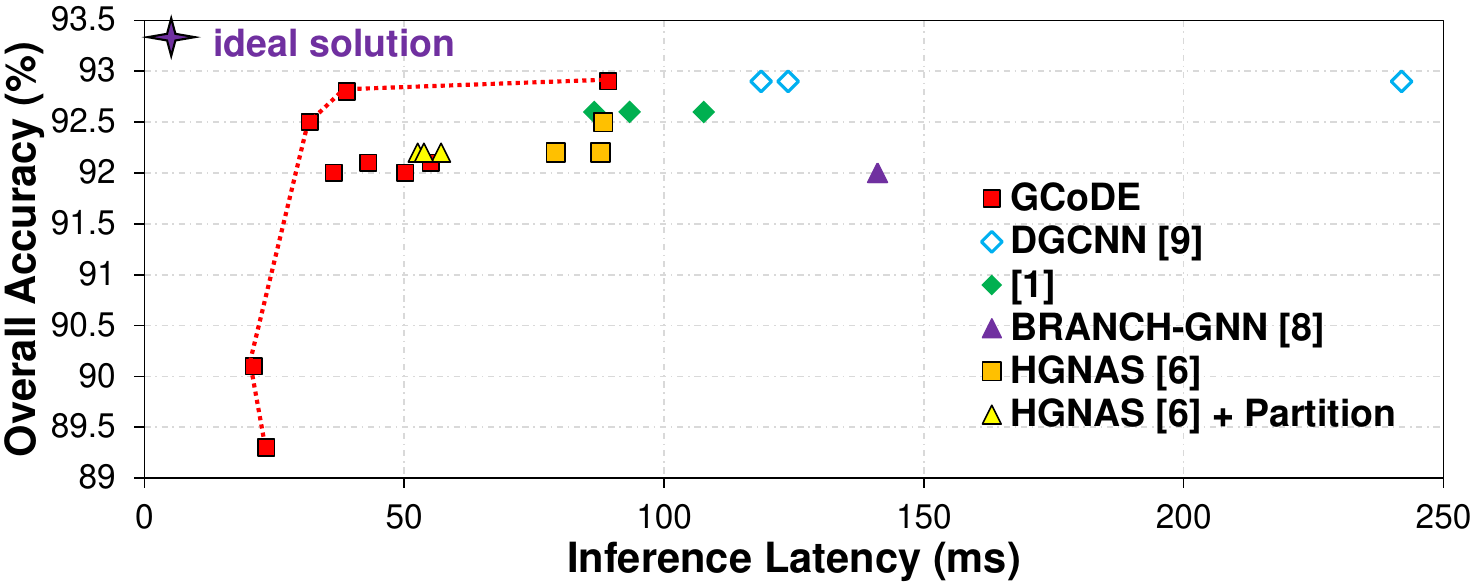}
    \caption{Comparison between the existing approaches and GCoDE in terms of accuracy and latency.}
    \label{fig:pareto}
    \vspace{-9pt}
\end{figure}

\textbf{Accuracy vs. Latency.}
Fig.~\ref{fig:pareto} presents the GNN design space exploration results using Jetson TX2 as the device.
Compared to all baselines, GCoDE can significantly push forward the Pareto frontier in GNN inference performance, achieving higher accuracy and lower latency.
This improvement is attributed to our proposed system performance predictor, which effectively identifies efficient GNN architectures, bringing GCoDE closer to the ideal solution.
Additionally, the selection of scaling factor $\lambda$ during the constraint-based random search process enables users to customize the GNN co-inference architecture for higher accuracy (smaller $\lambda$) or lower latency (larger $\lambda$), as needed.

\begin{table}[t]
\centering
\caption{Comparison of existing methods and GCoDE on MR.}
\renewcommand\arraystretch{1.0}
\resizebox{0.9\linewidth}{!}{%
\begin{tabular}{|ll|c|c|cc|c|}
\hline
\multicolumn{2}{|l|}{}                                           & \textbf{GCoDE} & BRANCHY \cite{shao2021branchy} & \multicolumn{2}{c|}{PNAS \cite{wei2023neural}} & \multicolumn{1}{l|}{PNAS \cite{wei2023neural}+Partition} \\ \hline
\multicolumn{2}{|c|}{Accuracy (\%)}                             & \textbf{76.1$\sim$77.0} & 75.5 & \multicolumn{2}{c|}{76.7} & 76.7  \\ \hline
\multicolumn{2}{|c|}{Mode}                                      & \textbf{Co}          & Co   & D           & E           & Co    \\ \hline
\multicolumn{1}{|l|}{\multirow{2}{*}{TX2 $\rightleftharpoons$ 1060}} & Latency (ms) & \textbf{8.70}   & 26.38  & 29.10        & 30.70        & 16.21                               \\
\multicolumn{1}{|l|}{}                           & Energy (J)   & \textbf{0.08}        & 0.26 & 0.31        & 0.28        & 0.17  \\ \hline
\multicolumn{1}{|l|}{\multirow{2}{*}{TX2 $\rightleftharpoons$ i7}}  & Latency (ms) & \textbf{8.50}         & 29.00   & 29.10        & 18.60       & 15.52 \\
\multicolumn{1}{|l|}{}                           & Energy (J)   & \textbf{0.08}        & 0.28 & 0.31        & 0.19        & 0.16  \\ \hline
\multicolumn{1}{|l|}{\multirow{2}{*}{Pi $\rightleftharpoons$ 1060}} & Latency (ms) & \textbf{4.80}         & 32.30 & 13.60        & 32.00          & 7.96  \\
\multicolumn{1}{|l|}{}                           & Energy (J)   & \textbf{0.03}        & 0.15 & 0.07        & 0.14        & 0.04  \\ \hline
\multicolumn{1}{|l|}{\multirow{2}{*}{Pi $\rightleftharpoons$ i7}}   & Latency (ms) & \textbf{2.00}           & 28.70 & 13.60        & 28.70        & 6.90  \\
\multicolumn{1}{|l|}{}                           & Energy (J)   & \textbf{0.01}        & 0.14 & 0.07        & 0.11        & 0.04  \\ \hline
\end{tabular}%
}
\label{tab:mr}
\vspace{-3pt}
\end{table}

\subsection{Evaluation on MR}

Tab.~\ref{tab:mr} reports the evaluation results on the MR dataset with high-dimensional node features, under a $40$ Mbps network condition, where GCoDE consistently maintains accuracy and efficiency.
Compared to the lightweight GNNs designed by PNAS, GCoDE achieves speedups of $3.3\times$, $2.2\times$, $2.8\times$, and $6.8\times$ in four heterogeneous system configurations, respectively.
Against co-inference approaches like BRANCHY and PNAS with optimal partitioning points, GCoDE achieves up to $14.3\times$ speedup.
Moreover, GCoDE stands out as the most energy-efficient approach compared to all baselines, requiring only $0.01$ J on the Raspberry Pi for a single inference.
These results strongly demonstrate the effective adaptation of GCoDE to system heterogeneity and its successful balance between communication and computation.

\subsection{Evaluation on System Performance Awareness}\label{sec:predictorResult}

\begin{figure}[t]
    \centering
    \includegraphics[width = 1\linewidth]{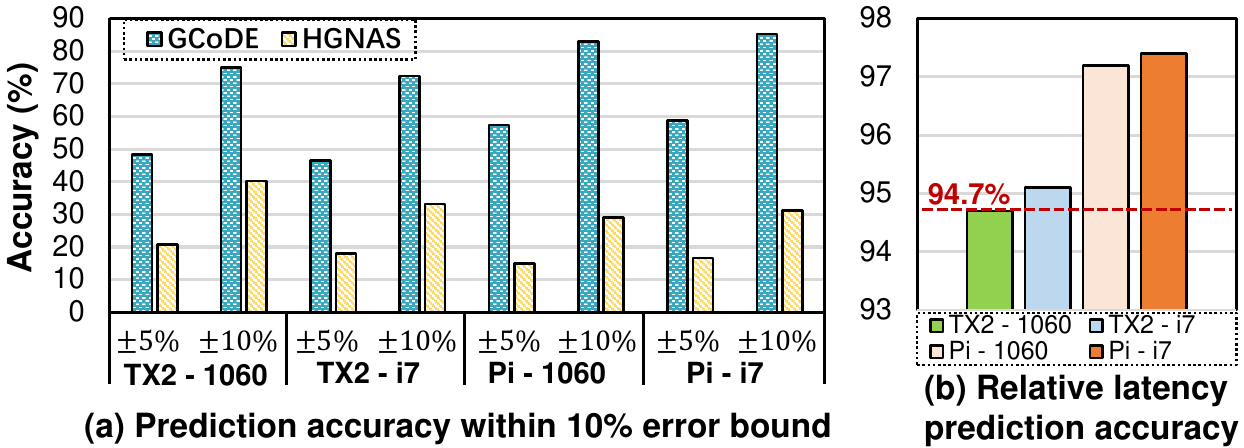}
    \caption{Latency prediction accuracy in co-inference systems.}
    \label{fig:predictor_result}
    \vspace{-3pt}
\end{figure}
\begin{sloppypar}

Fig.~\ref{fig:predictor_result}(a) shows that the proposed system performance predictor achieves $72.4\%$ to $85.3\%$ latency prediction accuracy across various system configurations within a $10\%$ error bound.
The high accuracy is attributed to the node feature enhancement approach, which considerably enhances the capacity of the predictor to discern hardware sensitivities in heterogeneous devices.
In contrast, HGNAS uses a one-hot encoding strategy for node feature initialization, leading to subpar prediction accuracy due to the lack of system heterogeneity information.
Furthermore, Fig.~\ref{fig:predictor_result}(b) demonstrates that GCoDE accurately evaluates the relative latency relationship among candidate architectures, achieving more than $94.7\%$ accuracy.
The effective capture of the relative latency relationship enables GCoDE to identify the more efficient architectures during the exploration.
Besides, our proposed cost estimation method, which requires no training overheads, attains over $88\%$ accuracy in predicting relative latency.
The results demonstrate that system performance awareness approaches of GCoDE effectively meet various application requirements and facilitate the exploration of efficient GNN co-inference designs.
\end{sloppypar}

\begin{figure}[t]
    \centering
    \includegraphics[width = 1\linewidth]{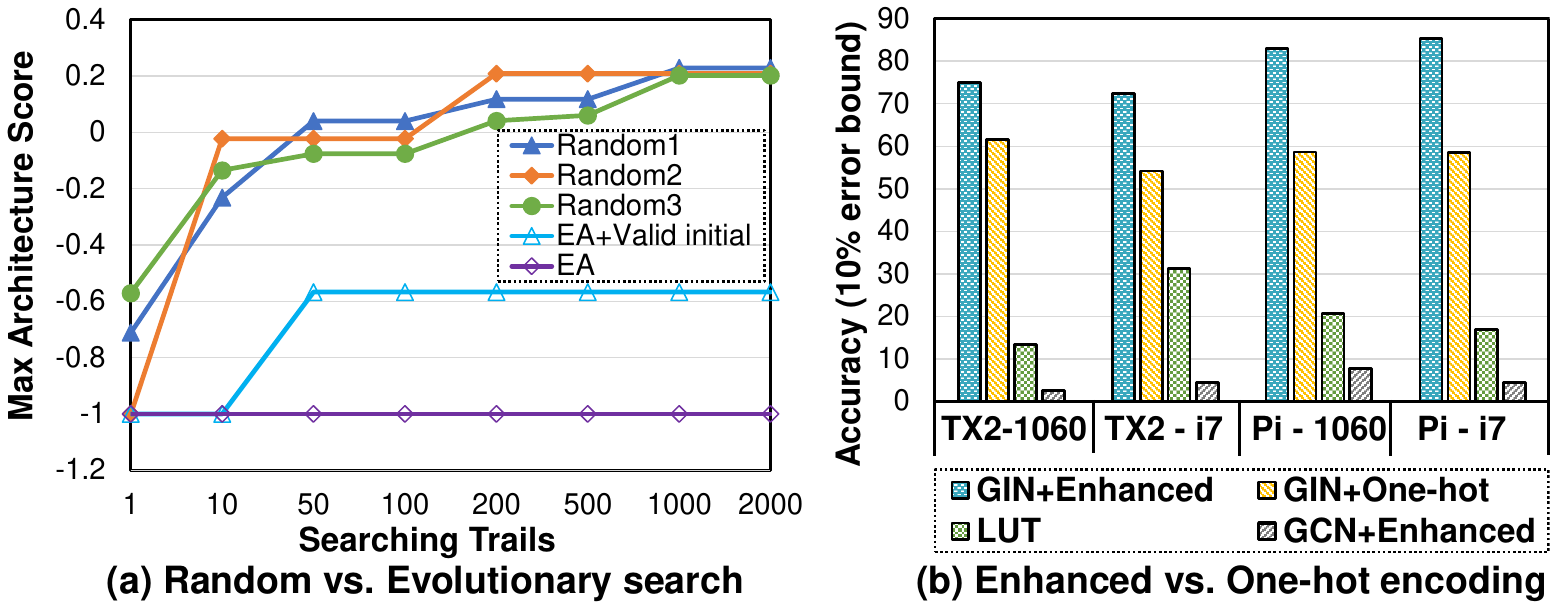}
    \caption{(a) Efficiency comparison between random- and evolutionary search. (b) Prediction accuracy improvement with GCoDE method (GIN + Enhanced feature).}
    \label{fig:ablation}
    \vspace{-5pt}
\end{figure}

\subsection{Ablation Studies}\label{sec:ablation}

\textbf{Random vs. Evolutionary search.}
Fig.~\ref{fig:ablation}(a) compares the search efficiency of random and evolutionary approaches. 
The results show that the EA gets stuck in a cycle of identifying valid architectures, thus missing chances to find higher-performing ones.
Even with an initial population of valid architectures, its performance remains sub-optimal.
In contrast, the proposed constraint-based random search strategy excels, finding optimal architectures within $2000$ trails (less than $3$ GPU hours).
Moreover, it also provides more freedom for GCoDE to explore several optimal solutions during a single search, enabling the building of GNN architecture zoo without additional overheads.

\textbf{Enhanced vs. One-hot encoding strategy.}
Fig.~\ref{fig:ablation}(b) shows that while the LUT method maintains accuracy in perceiving relative latency relationships, it is far from the truth latency value.
Additionally, GIN outperforms GCN in latency learning for architecture graphs due to its superior graph information learning capability.
Moreover, with the powerful graph information extraction capability of GIN, the one-hot encoding strategy obtains an improvement in accuracy, but is still ineffective.
In contrast, our feature enhancement method significantly improves prediction accuracy in diverse heterogeneous co-inference systems.


\subsection{Insight from GNNs Designed by GCoDE}

\begin{figure}[t]
    \centering
    \includegraphics[width = 1\linewidth]{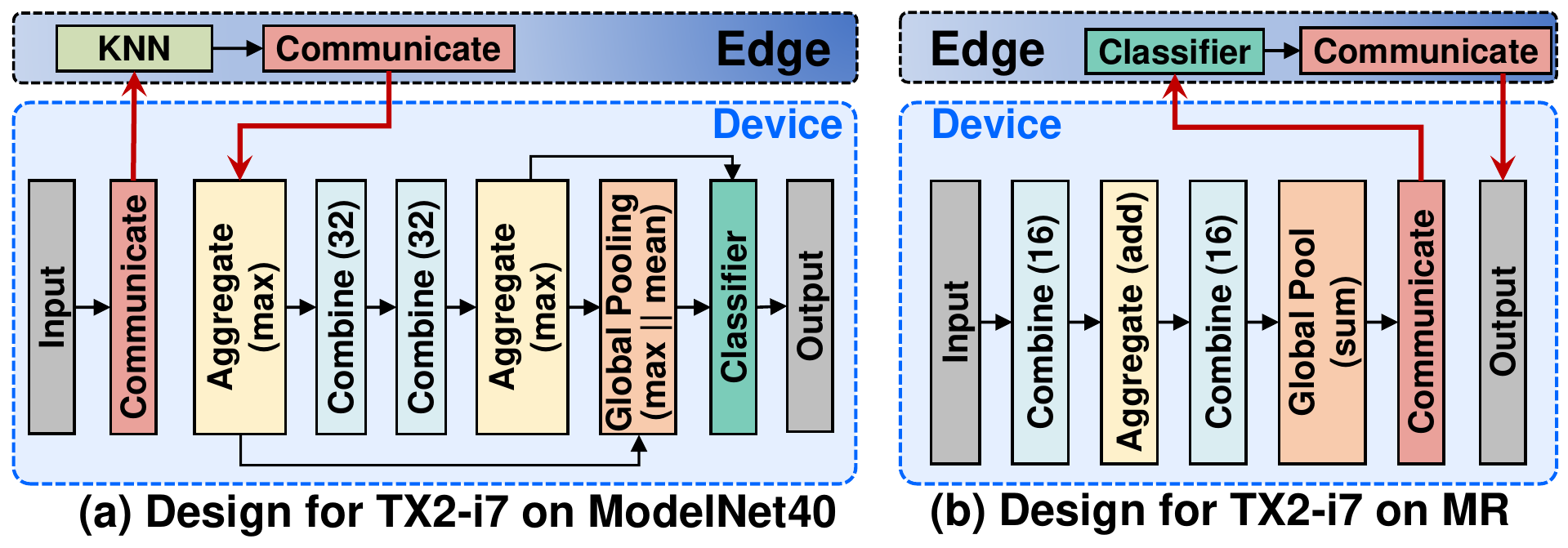}
    \caption{Visualization of GNNs designed by GCoDE.}
    \label{fig:insight}
    \vspace{-8pt}
\end{figure}

Fig.~\ref{fig:insight} visualizes the GNN designs by GCoDE tailored for the TX2-i7 co-inference system.
The results clearly show that GNNs designed by GCoDE with system-aware and architecture-mapping co-optimization, align effectively with heterogeneous hardware characteristics and balance communication-computation trade-offs, mirroring the observations in the \textbf{Motivation} section.
For the ModelNet40 dataset, GCoDE maps \textit{KNN} operation, which is inefficient on Jetson TX2, to the \textit{KNN}-friendly Intel i7, achieving improved performance with minimal communication costs (low feature dimension).
For the MR dataset, GCoDE allocates \textit{Combine} operation, which is a bottleneck on Intel i7, to Jetson TX2 and transfers the intermediate data to Intel i7 after feature dimension reduction by \textit{Global Pooling}.
Additionally, all designed architectures by GCoDE have been significantly simplified by eliminating redundant operations and function scale-down tuning.
Moreover, the co-inference engine executes operations on the device and edge in a pipelined manner, effectively mitigating communication overhead.

\section{Conclusions}

This paper introduces GCoDE, the pioneering system-aware automated framework for designing and deploying GNNs on device-edge co-inference systems.
GCoDE can automatically design tailored GNN architectures and deploy them on the target system using an efficient co-inference engine and runtime dispatcher.
GCoDE builds a unified architecture-mapping co-design space, leveraging constraint-based search strategies and accurate system performance awareness approaches to identify optimal solutions.
Extensive experiments show that GCoDE achieves superior accuracy, inference speed, and energy efficiency across diverse applications and systems, surpassing current SOTA methods with up to $44.9\times$ speedup and $98.2\%$ energy savings.
We believe that GCoDE has made an important heuristic step towards the design and deployment of efficient GNNs for large-scale wireless network edge applications.

\bibliographystyle{unsrt}

\bibliography{ref}

\end{document}